\documentclass[runningheads]{llncs}

\usepackage{eccv}

\usepackage{eccvabbrv}

\usepackage{graphicx}
\usepackage{booktabs}
\usepackage{amsmath}
\usepackage{newtxmath}
\usepackage{fontawesome} 
\usepackage[accsupp]{axessibility}  
\usepackage{pgfplots}
\usepgfplotslibrary{groupplots}

\usepackage[pagebackref,breaklinks,colorlinks]{hyperref}

\usepackage{orcidlink}

\begin{document}

\title{Caption-Driven Explorations: Aligning Image and Text Embeddings through Human-Inspired Foveated Vision} 

\titlerunning{NevaClip: Zero-shot Modeling of Human Attention in Captioning Tasks} 

\author{Dario Zanca \inst{1,*}\orcidlink{0000-1111-2222-3333} \and
Andrea Zugarini\inst{2}\orcidlink{0000-0002-9133-8669} \and
Simon Dietz\inst{1}\orcidlink{2222--3333-4444-5555} \and Thomas R. Altstidl\inst{1}\orcidlink{2222--3333-4444-5556} \and
Mark A. Turban Ndjeuha\inst{1}\orcidlink{2222--3333-4444-5557} \and Leo Schwinn\inst{3}\orcidlink{2222--3333-4444-5558} \and 
Bjoern M. Eskofier\inst{1,4}\orcidlink{2222--3333-4444-5559}
}
\authorrunning{D. Zanca et al.}

\institute{FAU Erlangen-Nürnberg
91052 Erlangen, Germany
\and
expert.ai, Siena, Italy
\and
Technical University of Munich, Munich, Germany
\and
Institute of AI for Health, Helmholtz Zentrum München, Munich, Germany\\
\textbf{ * \faEnvelope{} dario.zanca@fau.de}
}

\maketitle
\setlength{\tabcolsep}{4pt}

\renewcommand{\thefootnote}{} 
\footnotetext{This manuscript has been accepted at the Human-inspired Computer Vision (HCV) ECCV 2024 Workshop as an extended abstract. A long version of the work can be found at \url{https://arxiv.org/abs/2305.12380}.}
\renewcommand{\thefootnote}{\arabic{footnote}} 

\begin{abstract}
Understanding human attention is crucial for vision science and AI. While many models exist for free-viewing, less is known about task-driven image exploration. To address this, we introduce CapMIT1003, a dataset with captions and click-contingent image explorations, to study human attention during the captioning task. We also present NevaClip, a zero-shot method for predicting visual scanpaths by combining CLIP models with NeVA algorithms. NevaClip generates fixations to align the representations of foveated visual stimuli and captions. The simulated scanpaths outperform existing human attention models in plausibility for captioning and free-viewing tasks. This research enhances the understanding of human attention and advances scanpath prediction models.

\end{abstract}

\section{Introduction} \label{sec:intro}

\begin{figure}[t]
    \centering
    \includegraphics[scale=0.75]{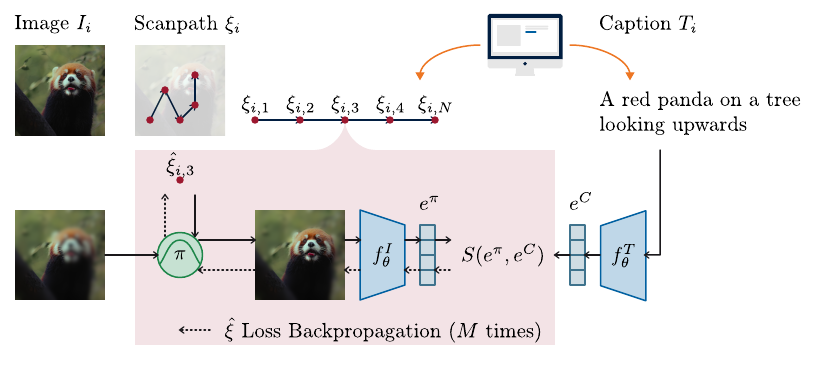}
    \caption{\textbf{Method Overview (NevaClip).} The third fixation $\xi_{i,3}$ is optimized in order to maximise the alignment between the caption representation $e^C$ and the foveated image representation $e^\pi$ under the current fixation point. }
    \label{fig:overview}
\end{figure}

\textbf{Motivations \& Related Work.} Visual attention is an essential cognitive process in human beings, enabling selective processing of relevant portions of visual input while disregarding irrelevant information \cite{kietzmann2011overt, macknik2009role}. Models of human attention are of great interest in neuroscience and applications \cite{cartella2024trends}, particularly for the case of scanpath prediction as it provides a more detailed understanding of visual attention dynamics compared to saliency prediction\cite{boccignone2019look,boccignone2019problems}. Seminal work  \cite{buswell1935people,tatler2010yarbus} investigated the relationship between eye movement patterns and high-level cognitive factors to demonstrate the influential role of task demands on eye movements and visual attention. Despite great interest in understanding mechanisms underlying task-driven visual exploration, most current classical \cite{itti1998model,boccignone2004modelling,le2007predicting,zanca2019gravitational} or machine learning-based \cite{pan2017salgan,wang2019revisiting,kummerer2022deepgaze} computational models focus on free viewing, which primarily involves bottom-up saliency, and overlooks the influence of different tasks. \\
\textbf{Contributions.} 
To investigate task-driven human attention and its interplays with language, we propose computational models that simulate human attention scanpaths during the captioning process.
First, we use a web application the expand the well-known MIT1003 dataset \cite{judd2009learning} with click-contingent task-driven image exploration. We call the new dataset CapMIT1003 and release it publicly.
Second, we combine Neural Visual Attention \cite{schwinn2022behind} algorithms with Contrastive Language-Image Pretrained (CLIP) \cite{radford2021learning} models, to generate task-driven scanpaths under human-inspired constraints of foveated vision.  
We found that generating scanpaths conditioned on the correctly associated captions results in highly plausible scanpath trajectories that achieve state-of-the-art results for the newly collected dataset.

\section{CapMIT1003 Dataset} 
We developed a web application that presents images from MIT1003 to participants and prompts them to provide captions while performing click-contingent image explorations, using the protocol defined in \cite{jiang2015salicon}. A click will reveal information in an area corresponding to two degrees of visual angle, to simulate foveated vision. Participants can click up to 10 times, before providing a caption. We instructed users to describe the content of the image with "one or two sentences", in English. All clicks and captions are stored in a database, while no personal information has been collected. In total, we collected 27865 clicks on 4573 observations over 153 sessions. We excluded 654 observations that were skipped, 33 with no recorded clicks, and 38 captions with less than 3 characters. The dataset is made publicly available \footnote{\url{https://huggingface.co/datasets/azugarini/CapMIT1003}}. 

\section{NevaClip Algorithm} 
An overview of the NevaClip algorithm is given in Figure \ref{fig:overview}. For each image $I_i$, a scanpath of user clicks $\xi_i$ and a caption $T_i$ are collected with a web interface. 

Let $\hat{I_i}$ be a blurred version of the image $I_i$. To predict the next fixation, i.e., $\xi_{i,t}$, with $t>0$, $t \in \mathbb{N}_0$ (non-negative integer), NevaClip's attention mechanism combines past predicted fixations $\{\hat{\xi}_{i,j} \mid j < t\}$ with an initial guess for $\hat{\xi}_{i,t}$ to create a foveated version of the image, 
\begin{align}\pi \left(I_i, t \right) =
\begin{cases}
\hat{I_i} & \text{if } t = 0 \\ 
\lambda \pi \left(I_i, t-1 \right) + \left[G_\sigma(\hat{\xi}_{i, t}) \cdot I_i + \left(1 - G_\sigma(\hat{\xi}_{i, t}) \cdot \hat{I_i} \right)\right] & \text{if } t > 0
\end{cases}
\end{align}
where $0<\lambda<1$ is a forgetting factor and $G_\sigma(\hat{\xi}_{i,t})$ is a Gaussian blob, with mean $\hat{\xi}_{i, t}$ and standard deviation $\sigma$. 
This image $\pi \left(I_i, t \right)$ is passed to CLIP's vision embedding network $f_\theta^I$ to obtain an image embedding $e^\pi$. The cosine similarity loss $S(e^\pi,e^C)$ between $e^\pi$ and the text embedding $e^C$ obtained by passing the caption to CLIP's text embedding network $f_\theta^T$ is then computed as a measure of how well the current scanpath aligns with the given caption. By backpropagating this loss $M$ times, and updating the value of $\hat{\xi}_{i, t}$ accordingly to minimize such loss, NevaClip finds the fixation $\hat{\xi}_{i,3}$ that minimizes this alignment loss. 

\section{Experiments}
\textbf{Experimental setup.} We adhere to the original implementation of the Neva algorithm \cite{schwinn2022behind} with 20 optimization steps and adapt it to run on CLIP backbones. Code is made publicly available \footnote{\url{https://github.com/dario-zanca/NevaClip}}. We use the implementation of the CLIP model provided by the original authors\footnote{\url{https://github.com/openai/CLIP}}. Unless stated otherwise, all NevaClip results are presented for the Resnet50 backbone.  For all  competitor models, the original code provided by the authors is used. 

\noindent\textbf{Baselines and models.} We evaluate three distinct variations of our NevaClip algorithm. For \textit{NevaClip (correct cap.)}, scanpaths are generated maximizing the alignment between visual embeddings of the foveated stimuli, and text embeddings of the corresponding captions. For \textit{NevaClip (different caption, same image) }, scanpath are generated to maximize the alignment between visual embeddings of the foveated stimuli, and text embeddings of the captions provided by a different subject on the same image. For \textit{NevaClip (different caption, different image)}, scanpath are generated to maximize the alignment between visual embeddings of the foveated stimuli, and text embeddings of the random captions provided by a different subject on a different image.
We include two baselines, i.e., Random and Center~\cite{judd2009learning}, to better position the quality of the results. We include also four competitor models~\cite{schwinn2022behind,boccignone2004modelling,zanca2019gravitational,koch1987shifts}.

\begin{table}[t]
\caption{\textbf{Average ScanPath Plausibility ($SPP$).} Summary of SPP SBTDE scores for all baselines and models. Best in \textbf{bold}.}
\centering
\begin{tabular}{lcc}
\hline
\textbf{} & \textbf{CapMIT1003} & \textbf{MIT1003}  \\
\textbf{Model} &  (Captioning) & (Free-viewing) \\
\hline
Random baseline & 0.26 (0.13) & 0.30 (0.11)  \\
Center baseline & 0.27 (0.13) & 0.33 (0.12) \\
\hline
G-Eymol~\cite{zanca2019gravitational} & 0.34 (0.17) & 0.46 (0.16) \\
Neva (original)~\cite{schwinn2022behind} & 0.34 (0.17)
 & 0.47 (0.16) \\
CLE~\cite{boccignone2004modelling} & 0.30 (0.19)
 & 0.47 (0.18) \\
WTA+Itti~\cite{koch1987shifts} & 0.31 (0.16) & 0.37 (0.14) \\
\hline
NevaClip (correct caption) & \textbf{0.35} (\textbf{0.18}) & \textbf{0.50} (\textbf{0.17}) \\
NevaClip (different caption, same image) & 0.34 (0.17) & \textbf{0.50} (\textbf{0.17}) \\
NevaClip (different caption, different image) & 0.26 (0.14) & 0.37 (0.15) \\
\hline
\end{tabular}
\label{tab:all_results}
\end{table}

\noindent \textbf{Scanpath similarity.}
To measure the similarity between simulated and human scanpaths, we compute ScanPath Plausibility (SPP)~\cite{fahimi2021metrics}, using the string-based time-delay embeddings (SBTDE)~\cite{schwinn2022behind} as a basic metric. In table \ref{tab:all_results} we summarise the SPP SBTDE scores for all baselines, competitors, and NevaCLIP versions. The scores are computed for each sublength $s \in \{1, ..., 10\}$, and the average of mean and standard deviation for all sublengths is presented. 

For CapMIT1003 (captioning task), \textit{NevaClip (correct caption)} outperforms all other approaches. As expected, \textit{NevaClip (different subject, different image)}, which generates a scanpath using a label from a different image and a different subject, performs similarly to the Random Baseline. The \textit{NevaClip (correct caption)}  performs slightly better than the \textit{NevaClip (different subject, same image)}, demonstrating that the caption provided by the subject brings useful information about their own visual exploration behavior. Among competitors, \textit{G-Eymol} and \textit{Neva (original)} compete well, despite not incorporating any caption information. It is worth noting that all approaches do not use any human gaze data for training, and the scanpath prediction is performed zero-shot. 

In MIT1003 (free-viewing), \textit{NevaClip (correct caption)} and \textit{NevaClip (different subject, same image)} also perform better than state-of-the-art models,
thus proving a substantial overlap between the two tasks. 

\section{Conclusion} 

As expected, capturing attention in captioning tasks proves to be more challenging compared to free-viewing, as it involves the simultaneous consideration of multiple factors such as image content, language, and semantic coherence. This is reflected in the results, where all approaches achieve lower performance in terms of scanpath plausibility (see table \ref{tab:all_results}) for the captioning task. The ranking of models remains consistent across both tasks, with free-viewing models performing reasonably well on both datasets. This observation suggests a substantial overlap between free viewing and captioning, possibly due to the initial phases of exploration being predominantly driven by bottom-up factors rather than task-specific demands.

A possible limitation of our work is represented by the click-contingent data collection. Future studies may complement this data collection with actual gaze data and explore differences between the two modalities.

\bibliographystyle{splncs04}
\bibliography{egbib}

\begin{thebibliography}{10}
\providecommand{\url}[1]{\texttt{#1}}
\providecommand{\urlprefix}{URL }
\providecommand{\doi}[1]{https://doi.org/#1}

\bibitem{boccignone2019look}
Boccignone, G., Cuculo, V., D’Amelio, A.: How to look next? a data-driven approach for scanpath prediction. In: International Symposium on Formal Methods. pp. 131--145. Springer (2019)

\bibitem{boccignone2019problems}
Boccignone, G., Cuculo, V., D’Amelio, A.: Problems with saliency maps. In: International Conference on Image Analysis and Processing. pp. 35--46. Springer (2019)

\bibitem{boccignone2004modelling}
Boccignone, G., Ferraro, M.: Modelling gaze shift as a constrained random walk. Physica A: Statistical Mechanics and its Applications  \textbf{331}(1-2),  207--218 (2004)

\bibitem{buswell1935people}
Buswell, G.T.: How people look at pictures: a study of the psychology and perception in art.  (1935)

\bibitem{cartella2024trends}
Cartella, G., Cornia, M., Cuculo, V., D'Amelio, A., Zanca, D., Boccignone, G., Cucchiara, R., et~al.: Trends, applications, and challenges in human attention modelling. In: Proceedings of the International Joint Conference on Artificial Intelligence (2024)

\bibitem{fahimi2021metrics}
Fahimi, R., Bruce, N.D.: On metrics for measuring scanpath similarity. Behavior Research Methods  \textbf{53}(2),  609--628 (2021)

\bibitem{itti1998model}
Itti, L., Koch, C., Niebur, E.: A model of saliency-based visual attention for rapid scene analysis. IEEE Transactions on pattern analysis and machine intelligence  \textbf{20}(11),  1254--1259 (1998)

\bibitem{jiang2015salicon}
Jiang, M., Huang, S., Duan, J., Zhao, Q.: Salicon: Saliency in context. In: Proceedings of the IEEE conference on computer vision and pattern recognition. pp. 1072--1080 (2015)

\bibitem{judd2009learning}
Judd, T., Ehinger, K., Durand, F., Torralba, A.: Learning to predict where humans look. In: 2009 IEEE 12th international conference on computer vision. pp. 2106--2113. IEEE (2009)

\bibitem{kietzmann2011overt}
Kietzmann, T.C., Geuter, S., K{\"o}nig, P.: Overt visual attention as a causal factor of perceptual awareness. PloS one  \textbf{6}(7),  e22614 (2011)

\bibitem{koch1987shifts}
Koch, C., Ullman, S.: Shifts in selective visual attention: towards the underlying neural circuitry. In: Matters of intelligence, pp. 115--141. Springer (1987)

\bibitem{kummerer2022deepgaze}
K{\"u}mmerer, M., Bethge, M., Wallis, T.S.: Deepgaze iii: Modeling free-viewing human scanpaths with deep learning. Journal of Vision  \textbf{22}(5), ~7--7 (2022)

\bibitem{le2007predicting}
Le~Meur, O., Le~Callet, P., Barba, D.: Predicting visual fixations on video based on low-level visual features. Vision research  \textbf{47}(19),  2483--2498 (2007)

\bibitem{macknik2009role}
Macknik, S.L., Martinez-Conde, S.: The role of feedback in visual attention and awareness. Cognitive Neurosciences  \textbf{1} (2009)

\bibitem{pan2017salgan}
Pan, J., Ferrer, C.C., McGuinness, K., O'Connor, N.E., Torres, J., Sayrol, E., Giro-i Nieto, X.: Salgan: Visual saliency prediction with generative adversarial networks. arXiv preprint arXiv:1701.01081  (2017)

\bibitem{radford2021learning}
Radford, A., Kim, J.W., Hallacy, C., Ramesh, A., Goh, G., Agarwal, S., Sastry, G., Askell, A., Mishkin, P., Clark, J., et~al.: Learning transferable visual models from natural language supervision. In: International conference on machine learning. pp. 8748--8763. PMLR (2021)

\bibitem{schwinn2022behind}
Schwinn, L., Precup, D., Eskofier, B., Zanca, D.: Behind the machine's gaze: Neural networks with biologically-inspired constraints exhibit human-like visual attention. arXiv preprint arXiv:2204.09093  (2022)

\bibitem{tatler2010yarbus}
Tatler, B.W., Wade, N.J., Kwan, H., Findlay, J.M., Velichkovsky, B.M.: Yarbus, eye movements, and vision. i-Perception  \textbf{1}(1),  7--27 (2010)

\bibitem{wang2019revisiting}
Wang, W., Shen, J., Xie, J., Cheng, M.M., Ling, H., Borji, A.: Revisiting video saliency prediction in the deep learning era. IEEE transactions on pattern analysis and machine intelligence  \textbf{43}(1),  220--237 (2019)

\bibitem{zanca2019gravitational}
Zanca, D., Melacci, S., Gori, M.: Gravitational laws of focus of attention. IEEE transactions on pattern analysis and machine intelligence  \textbf{42}(12),  2983--2995 (2019)

\end{thebibliography}
\end{document}